\documentclass[12pt]{article}
\usepackage[utf8]{inputenc}
\usepackage[T1]{fontenc}
\usepackage{amssymb}
\usepackage{amsmath}
\usepackage{amsfonts}
\usepackage{stfloats}
\usepackage [autostyle, english = american]{csquotes}
\MakeOuterQuote{"}
\usepackage{subcaption}
\usepackage{textcomp}
\usepackage{blindtext}
\usepackage{graphicx}
\usepackage{lettrine}
\usepackage{lipsum}
\usepackage{float}
\usepackage{multicol}
\usepackage{multirow}
\usepackage{textcomp}
\usepackage{layouts}
\usepackage[font=small,skip=0pt]{caption}
\usepackage{color}
\usepackage[table]{xcolor}
\usepackage{mathtools}
\usepackage{booktabs}

\usepackage{array}
\newcolumntype{L}[1]{>{\raggedright\let\newline\\\arraybackslash\hspace{0pt}}m{#1}}
\newcolumntype{C}[1]{>{\centering\let\newline\\\arraybackslash\hspace{0pt}}m{#1}}
\newcolumntype{R}[1]{>{\raggedleft\let\newline\\\arraybackslash\hspace{0pt}}m{#1}}
\usepackage[linesnumbered]{algorithm2e}
\usepackage{bbm}
\usepackage{pgf-pie}
\usepackage{tikz}
\usepackage{duckuments}
\usepackage[hidelinks]{hyperref}
\usepackage[left=2cm, right=3cm, top=2cm, bottom=1cm]{geometry}
\usepackage{xcolor}
\definecolor{myred}{rgb}{0.7529,0,0}  
\usepackage{fullpage}
\usepackage{pgf-pie}
\usepackage{pgfplots}
\usepackage{rotating}
\usepackage{booktabs}
\usepackage{longtable}
\usepackage{tikz}
\usepackage{graphicx}
\usepackage{epstopdf}
\usepackage{authblk}
\usepackage{epstopdf}
\usepackage{etoolbox}
\usetikzlibrary{shapes.geometric, arrows}
\usetikzlibrary{positioning}
\usepackage[shortlabels]{enumitem}
\usepackage{pdflscape}
\usepackage{rotating}
\begin{document}

\title{Neural Moving Horizon Estimation:\\ A Systematic Literature Review}

\author[1]{Surrayya Mobeen}
\author[1]{Jann Cristobal}
\author[1]{Shashank Singoji}
\author[1]{Basaam Rassas}
\author[1]{Mohammadreza Izadi}
\author[1]{Zeinab Shayan}
\author[1]{Amin Yazdanshenas}
\author[1]{Harneet Kaur}
\author[1]{Robert Barnsley}
\author[1]{Lana Elliott}
\author[1,*]{Reza Faieghi}
\affil[1]{\small{Autonomous Vehicles Laboratory, Department of Aerospace Engineering, Toronto Metropolitan University, 350 Victoria St., M5B2K3, Toronto, Ontario, Canada}}
\affil[*]{Corresponding author: Email: reza.faieghi@torontomu.ca}
\date{}
\maketitle
\begin{abstract}
The neural moving horizon estimator (NMHE) is a relatively new and powerful state estimator that combines the strengths of neural networks (NNs) and model-based state estimation techniques. 
Various approaches exist for constructing NMHEs, each with its unique advantages and limitations.
However, a comprehensive literature review that consolidates existing knowledge, outlines design guidelines and highlights future research directions is currently lacking.
This systematic literature review synthesizes the existing knowledge on NMHE, addressing the above knowledge gap.
The paper (1) explains the fundamental principles of NMHE, (2) explores different NMHE architectures, discussing the pros and cons of each, (3) investigates the NN architectures used in NMHE, providing insights for future designs, (4) examines the real-time implementability of current approaches, offering recommendations for practical applications, and (5) discusses the current limitations of NMHE approaches and outlines directions for future research. 
These insights can significantly improve the design and application of NMHE, which is critical for enhancing state estimation in complex systems.\\

\textbf{Keywords:}
Moving horizon estimation, neural network, state estimation
\end{abstract}



\section{Introduction}
State estimation is a fundamental problem in control engineering, crucial for ensuring the monitoring and control of dynamic systems. Achieving accurate and robust state estimation can be a formidable challenge due to sensor noise, incomplete or indirect measurements, nonlinearities, model uncertainties, and external disturbances \cite{simon2006optimal}.

Conventional state estimators, such as Kalman filters \cite{khodarahmi2023review}, particle filters \cite{doucet2009tutorial}, and moving horizon estimators (MHEs) \cite{rawlings2021moving}, rely on a model of the system; hence the term model-based. These methods, while grounded in strong theoretical foundations and capable of providing precise estimates when accurate models are available, often struggle with handling model inaccuracies and computational complexity, particularly in nonlinear and time-varying systems. On the contrary, NN-based state estimation techniques offer greater flexibility and adaptability by learning directly from data. These methods can effectively manage complex and nonlinear systems without requiring explicit mathematical models. Still, their performance heavily depends on the quality and quantity of the training data, and they often lack transparency.

Recognizing the limitations of both model-based and NN-based approaches, there has been growing interest in hybrid methods that aim to combine the strengths of both \cite{s21062085}. By integrating the predictive accuracy and theoretical robustness of model-based techniques with the flexibility and adaptability of NN-based methods, hybrid approaches can provide more robust and accurate state estimation solutions, addressing many of the challenges inherent in traditional state estimation methods. Reviews of existing approaches that integrate Kalman filters and NNs are available in \cite{feng2023review, bai2023state}.

One growing class of hybrid state estimation algorithms is rooted in the integration of MHE and NNs. MHE is a powerful model-based state estimation technique that uses a finite window of recent measurements to solve a constrained optimization problem for state estimation. The optimization problem aims to minimize the difference between expected and measured outputs over a moving horizon \cite{inbook}.

MHE is closely related to model predictive control (MPC), where future system behavior is predicted based on the current state estimate, and a control sequence is optimized while accounting for system constraints \cite{schwenzer2021review}. Although MHE may not be as widely known as MPC, its similarities to MPC offer certain advantages for state estimation. 

The ability of MHE to use a sequence of past measurements and optimize state estimates with respect to system dynamics, control inputs, and measurement noise makes it highly effective in handling nonlinear and time-varying dynamics that could be subject to input and state constraints as well as measurement noise and faults \cite{FAGIANO2013193, wan2018real}. As such, MHE is employed in a wide range of applications, including process control \cite{tenny2002efficient}, robotics \cite{bae2017humanoid}, self-driving cars \cite{zanon2013nonlinear}, and aerospace systems \cite{vandersteen2013spacecraft}, demonstrating its versatility and effectiveness in various fields.

However, MHE's performance heavily relies on the accuracy of the system model \cite{kraus2013moving}. Integrating NNs with MHE can help mitigate this issue. NNs learn system dynamics from data, reducing dependence on precise mathematical models and improving robustness to model inaccuracies \cite{11}. They can also assist in the adaptive tuning of MHE parameters, optimizing performance over time. By leveraging the pattern recognition and approximation capabilities of NNs, the integration can lead to more accurate and reliable state estimation, addressing many of the inherent limitations of classical MHE \cite{7159360}.

There have been different approaches to integrating NNs with MHE, each presenting certain advantages. However, a comprehensive resource discussing the current Neural MHE (NMHE) designs is lacking. This motivated us to systematically review the existing literature, providing readers with a thorough grasp of the capabilities, constraints, and possible applications of NMHE. 
It is worth noting that other machine learning algorithms, such as decision trees and Gaussian process regression, have also been combined with MHE \cite{2, 22}. However, it is the NNs that are the most commonly used learning models with the most advantages in the context of MHE, and our focus is specifically on them.
That said, in this systematic review paper, we aim to answer the following questions:
\begin{itemize}
\item What are the different approaches to designing NMHE?
\item What are the different structures of NNs used in NMHE?
\item How do different techniques compare in terms of state estimation accuracy and computational efficiency?
\end{itemize}
The answers to the above questions clarify the pros and cons of existing approaches, generate recommendations for the design and implementation of NMHE for a given application, and identify the existing gaps and future research directions.

\section{Overview of NMHE}\label{se:MHEformulation}
Before we start with the systematic review, we present an overview of MHE and its relation with NN methods.
This clarifies the terminology and concepts, setting up the stage for our discussions in the subsequent sections.

\subsection{MHE formulation}
\begin{figure}[htbp]
    \centering
    \includegraphics[width=0.8\linewidth]{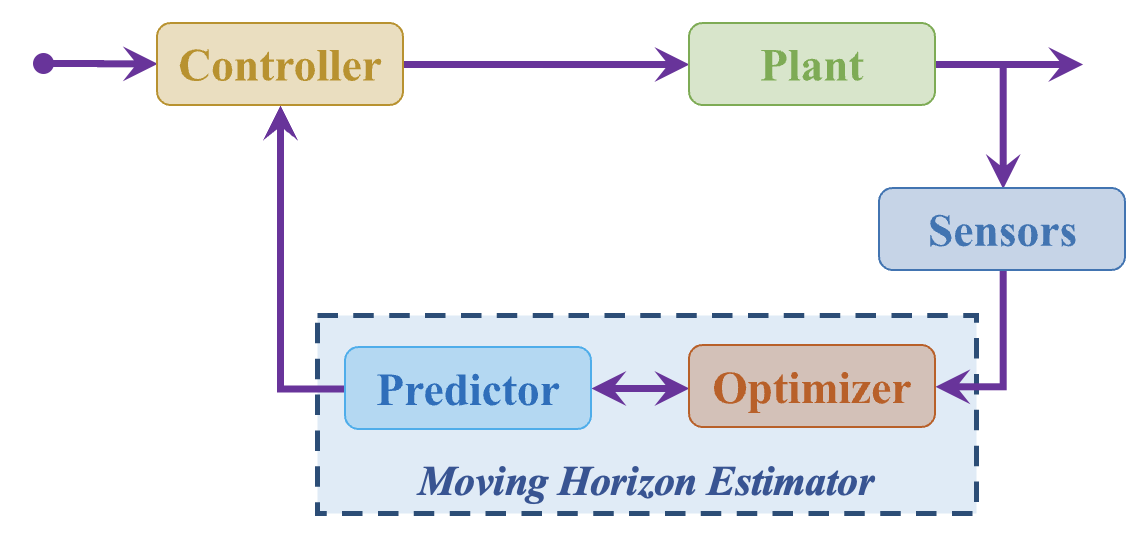}
    \caption{Block diagram of MHE within a close-loop control system}
    \label{fig:MHE block diagram}
\end{figure}

\begin{figure}[htbp]
    \centering
    \includegraphics[width=0.9\textwidth]{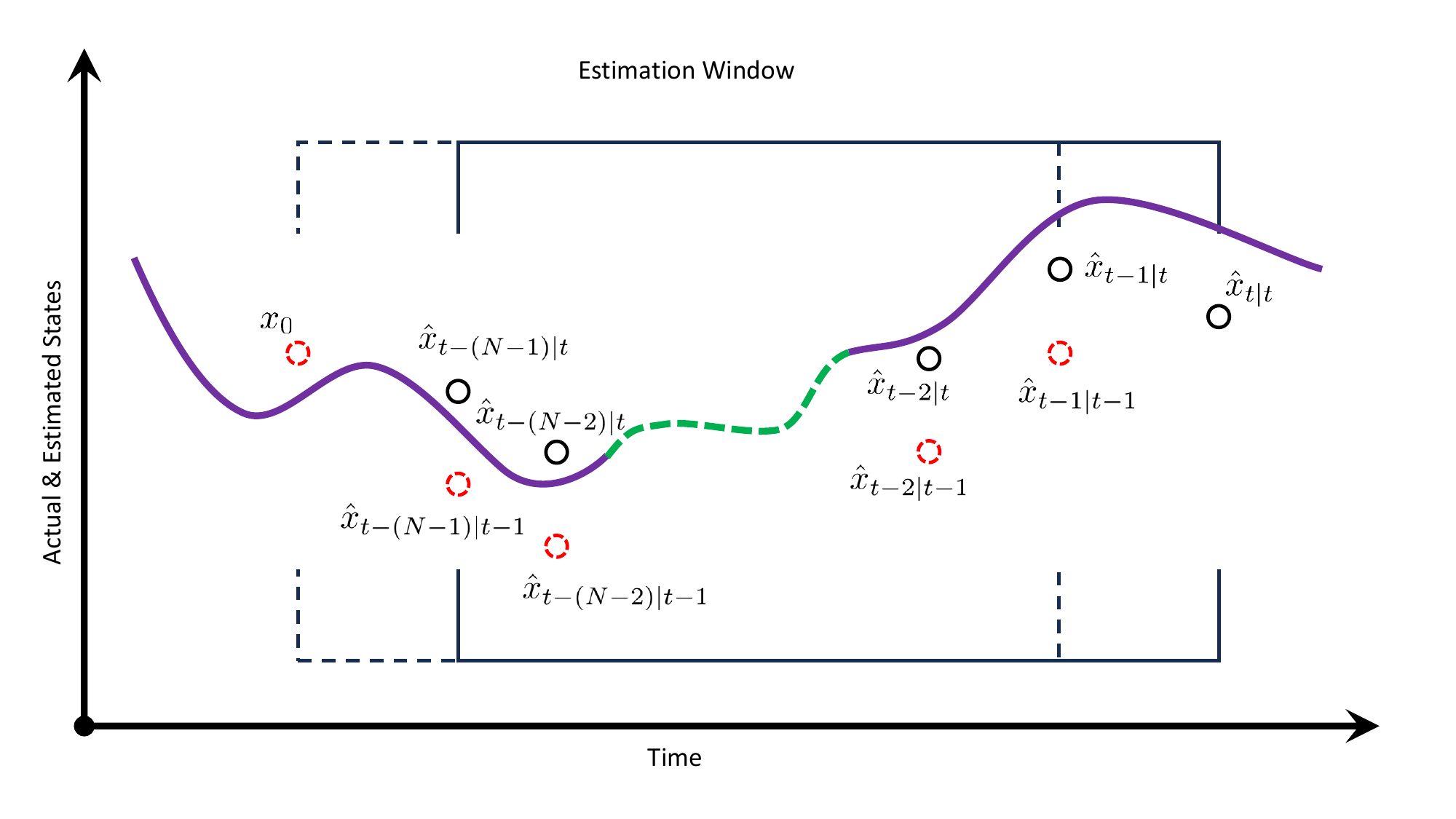}
    \caption{Illustration of moving horizon in MHE}
    \label{fig:mhehorizon}
\end{figure}

Figure \ref{fig:MHE block diagram} displays a basic architecture of MHE implemented in a closed-loop control system.
The main components of MHE are a predictor and an optimizer.
The predictor forecasts future states of the system based on the current state estimate and control input.
The optimizer refines the state estimates by solving a constrained optimization problem \cite{rawlings2021moving}.
Both predictor and optimizer use a sequence of previous measurements within a moving window.
Figure \ref{fig:mhehorizon} illustrates how the window shifts over time, incorporating past measurements and predictions to update and refine state estimations iteratively.
This iteration enables the MHE algorithm to continuously adjust its predictions based on new information, resulting in improved accuracy and reliability in estimating the underlying states of the system.

To establish the mathematical formulation of MHE, consider a discrete-time nonlinear system described by the following state-space equations:
\begin{equation}
\begin{array}{l}
     x_{k+1} = f\left(x_k, u_k\right) + w_k,\\
     y_k = h \left( x_k \right) + v_k,
\end{array}
\label{eq:system}
\end{equation}
where $x$ is the state vector, $u$ is the control input, $y$ is the measurement vector, $w$ is the process noise, $v$ is the measurement noise, $f\left(\cdot, \cdot \right)$ is the state transition function, and $h\left(\cdot \right)$ is the measurement function.
The MHE aims to estimate the state sequence $\{x_{k-N}, x_{k-N+1}, \ldots, x_k\}$ over a moving horizon of length $N$, given the measurements $\{y_{k-N}, y_{k-N+1}, \ldots, y_k\}$ and control inputs $\{u_{k-N}, u_{k-N+1}, \ldots, u_k\}$.

For the optimizer, the cost function typically includes terms for the deviation from the predicted state trajectory and the measurement error
\begin{equation}
J = \| x_{k-N} - \hat{x}_{k-N} \|_P^2 + \sum_{i=k-N}^{k-1} \left[ \| y_i - h(\hat{x}_i) \|_R^2 + \| x_{i+1} - f(\hat{x}_i, u_i) \|_Q^2 \right],
\label{eq:cost}
\end{equation}
where $R$ is the measurement noise covariance matrix, $Q$ is the process noise covariance matrix, $P$ is the covariance matrix associated with the initial state estimate $\hat{x}_{k-N}$, and $\|\cdot\|_R^2$ denotes the weighted squared norm defined as $\|a\|_R^2 = a^T R^{-1} a$.

Let us explore the terms within \eqref{eq:cost} in detail.
The initial state error term \(\| x_{k-N} - \hat{x}_{k-N} \|_P^2\) penalizes the difference between the initial state estimate \( x_{k-N} \) at the beginning of the horizon and a prior state estimate \( \hat{x}_{k-N} \). By minimizing this term, the estimator ensures that the state estimates at the start of the horizon window are consistent with prior knowledge or previous estimates, thereby providing a good starting point for the optimization.

The measurement error term  $\| y_i - h(\hat{x}_i) \|_R^2$ penalizes the difference between the actual measurements \(y_i\) and the predicted measurements \(h(\hat{x}_i)\) based on the current state estimate $\hat{x}_i$. By minimizing this term, the estimator ensures that the state estimates $\hat{x}_i$ are consistent with the observed measurements $y_i$, thereby improving the accuracy of the state estimation.

The state transition error term $\| x_{i+1} - f(\hat{x}_i, u_i) \|_Q^2$ penalizes the difference between the predicted next state $f(\hat{x}_i, u_i)$ and the actual next state $x_{i+1}$. By minimizing this term, the estimator ensures that the predicted state transitions are consistent with the actual state transitions, thereby maintaining the dynamic consistency of the state estimates.

By combining these terms in the cost function, MHE aims to strike a balance between accurately estimating the initial state, aligning the predicted and measured outputs, and maintaining consistency in the estimated states over time.

Note that for parameter estimation, one can add the weighted squared norms of terms like ${\theta}_{t-N} - \hat{\theta}_{t-N}$ to the cost function, where $\theta$ is a parameter to be estimated.

With the cost function established, the MHE problem formulation is
\begin{equation}
\label{eq:mhe_formulation}
\min_{\{x_{k-N}, x_{k-N+1}, \ldots, x_k\}} J,
\end{equation}
subject to
\begin{equation}
\begin{array}{ll}
   \text{state dynamics constraints:}  &  x_{i+1} = f(x_i, u_i) + w_i, \quad i = k-N, \ldots, k-1, \\
   \text{measurement constraints:}  & y_i = h(x_i) + v_i, \quad i = k-N, \ldots, k,\;\text{and} \\
   \text{initial state constraint:} & x_{k-N} = \hat{x}_{k-N} + w_{k-N}. \\
\end{array}
\end{equation}


\subsection{Integration of NNs with MHE}
One limitation of MHE is its reliance on an accurate system model, which can be challenging to obtain or may not adequately capture all system dynamics.
Recent advances in NNs have proven them to be powerful tools for learning complex relationships from data. 
Therefore, integrating NNs with MHE can significantly enhance MHE's capabilities.

To elaborate, NNs excel at capturing complex nonlinear relationships between input and output variables, making them well-suited for modeling dynamic systems within MHE formulation.
Their adaptability allows NNs to continuously learn from data, enabling robust state estimation even in the case of time-varying system dynamics.
Additionally, NNs can automatically extract relevant features from raw sensor data.
This eliminates the need for manual feature engineering, especially in control systems with multiple sensors where raw sensor data is high-dimensional and complex.
Moreover, NNs are well-suited for parallel computing, facilitating efficient computation necessary for real-time applications.

However, NNs also present certain limitations in MHE applications.
One notable challenge is the requirement for large amounts of training data, which may be difficult to obtain and can lead to reduced performance in cases of limited or noisy data.
Furthermore, the inherent complexity of NNs can hinder their interpretability, making it challenging to understand and validate the underlying estimation process.
Overfitting is another concern, where the model may memorize noise or spurious patterns in the training data, compromising its generalization performance on unseen data.
Additionally, deploying NMHE may require significant computational resources, posing constraints in resource-constrained environments or real-time applications.

Despite these limitations, the advantages of integrating NNs with MHE hold great promise for addressing complex state estimation problems.
Understanding and mitigating the limitations can help researchers harness the full potential of NMHE. 
This motivates a systematic review to study the existing approaches, generate design guidelines for future NMHE implementations, and inform existing gaps and future research directions. 
\section{Review Methodology}
To ensure a comprehensive review of the literature and enhance the rigor of our analysis, we followed the Preferred Reporting Items for Systematic reviews and Meta-Analyses (PRISMA) guidelines \cite{page2021prisma}.
This section presents the details of our review methodology.

\subsection{Search Strategy and Criteria}
Our search terms consisted of two key concepts of NNs and MHE.
Table \ref{tab:search terms} presents the controlled and uncontrolled terms for our search.
We searched three different databases, including IEEE Xplore, Scopus, and Google Scholar, for all studies published or accepted before 5 March 2024.

\begin{footnotesize}
\begin{longtable}{@{}p{\linewidth}@{}}
\caption{Search terms used within the Databases}
\label{tab:search terms} \\
\toprule  
\multicolumn{1}{c}{\textbf{Search terms include controlled and uncontrolled terms}}\\
\midrule
\textbf{Controlled Terms}: Neural Networks, Artificial Neural Networks, Moving Horizon Estimator, MHE \\
\midrule
\textbf{Uncontrolled Terms}: Deep Networks, Receding Horizon, NN Estimation, Deep Neural Networks \\
\bottomrule
\end{longtable}
\end{footnotesize}

\subsubsection{Inclusion and Exclusion Criteria}
We devised and applied the following inclusion and exclusion criteria to ensure a comprehensive and focused review of the literature on NMHE.

\begin{description}
    \item[Inclusion Criteria:]
    \begin{enumerate}
        \item Studies that directly involve NNs as a fundamental component of their data-driven MHE methodology.
        \item Peer-reviewed journal articles, conference papers, theses, and dissertations to ensure academic rigor.
        \item Research conducted within the last 15 years.
        \item Publications in the English language.
    \end{enumerate}
\end{description}

\begin{description}
    \item[Exclusion Criteria:]
    \begin{enumerate}
        \item Studies that do not focus on NMHE.
        \item Duplicate publications or multiple versions of the same study.
        \item Studies with limited relevance to the topic, such as those that only briefly mention NNs or provide minimal details about their application.
        \item Studies published before the specified time frame to ensure the focus on recent research.
        \item Studies that mention NNs but do not use them as a central component of their data-driven MHE approach.
        \item Publications in languages other than English, unless they provide English translations or summaries.
    \end{enumerate}
\end{description}

\subsection{Study Selection}
\begin{figure}[htbp]
    \centering
    \includegraphics[width=0.95\textwidth]{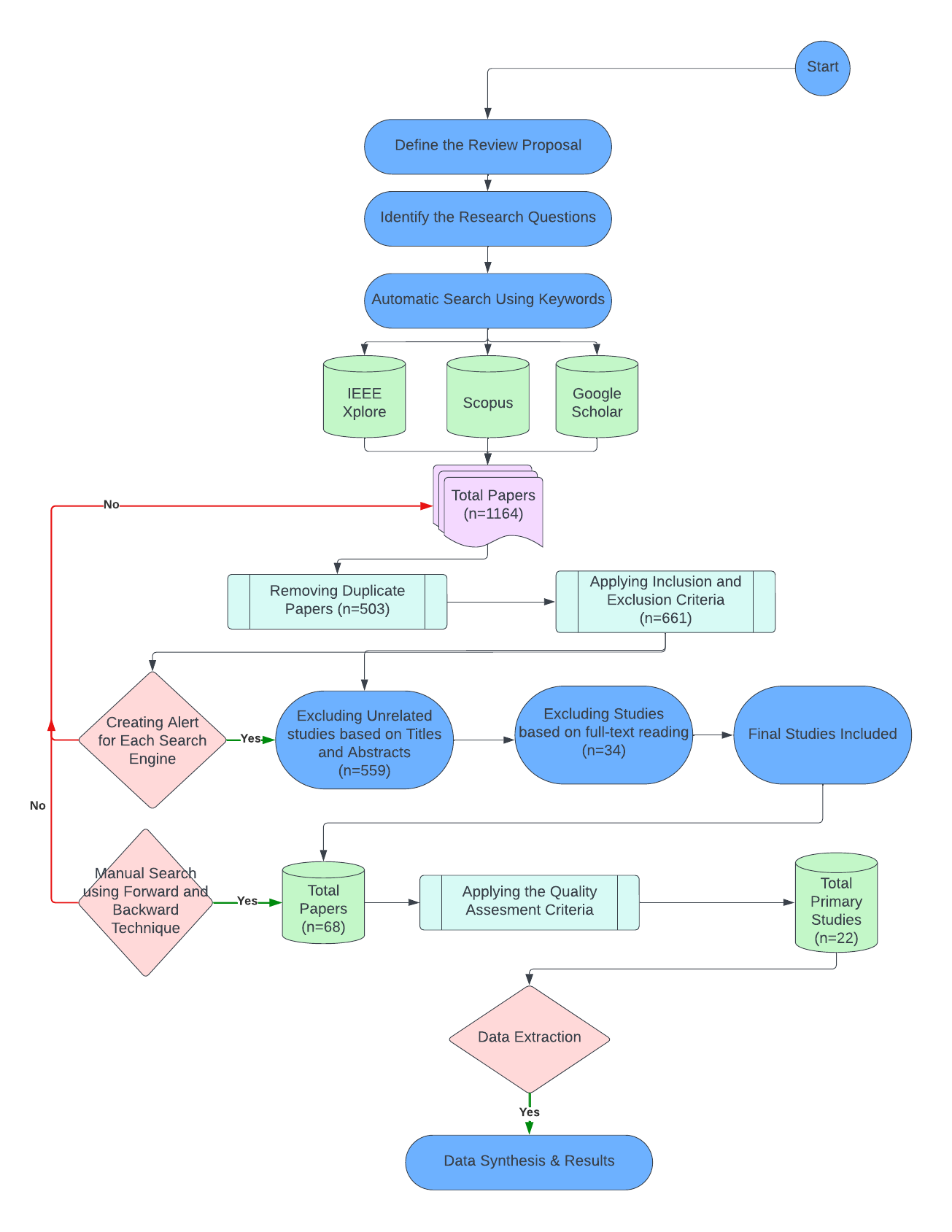}
    \caption{Overview of review methodology}
    \label{fig:flowchart}
\end{figure}

The search from the three databases yielded 1164 studies. 
We merged the search results, removing 503 duplicates.
We screened the studies to ensure relevance to NMHE.
This entailed an initial screening where two reviewers independently screened the titles and abstracts of the studies.
When conflicts arose, a third reviewer resolved them.
We repeated this procedure for the full-text screening, resulting in 69 studies. 
After quality assessment, 22 studies remained for data extraction. 
The quality assessment focused on the relevance of the studies.
Once the full-text screening was completed, team members who had not participated in either the initial or full-text screening assessed whether the studies addressed our research questions and the objectives of this literature review.
Figure \ref{fig:flowchart} details the step-by-step implementation of the search strategy.

\subsection{Data Extraction}
Our systematic literature review used a detailed set of data extraction questions to provide a robust framework for assessing and synthesizing research findings from various studies.
These questions covered key topics such as NN structures, algorithm effectiveness, experiment designs, computational efficiency of algorithms, comparative analysis with previous approaches, and conclusions.

Two reviewers independently extracted data from each study, ensuring a thorough and unbiased assessment.
When conflicts arose, a third reviewer resolved them to maintain consistency and accuracy.
By methodically summarizing data, we highlighted common themes, advantages, and disadvantages of different approaches, and other crucial insights.
Ultimately, this systematic data extraction process provided a strong foundation for drawing robust conclusions and making evidence-based decisions.

The following sections present the results of this systematic review and the lessons learned from it.
\section{Descriptive Statistics}\label{se:descriptive}
\begin{figure}[ht]
    \centering
    \includegraphics[width=0.9\textwidth]{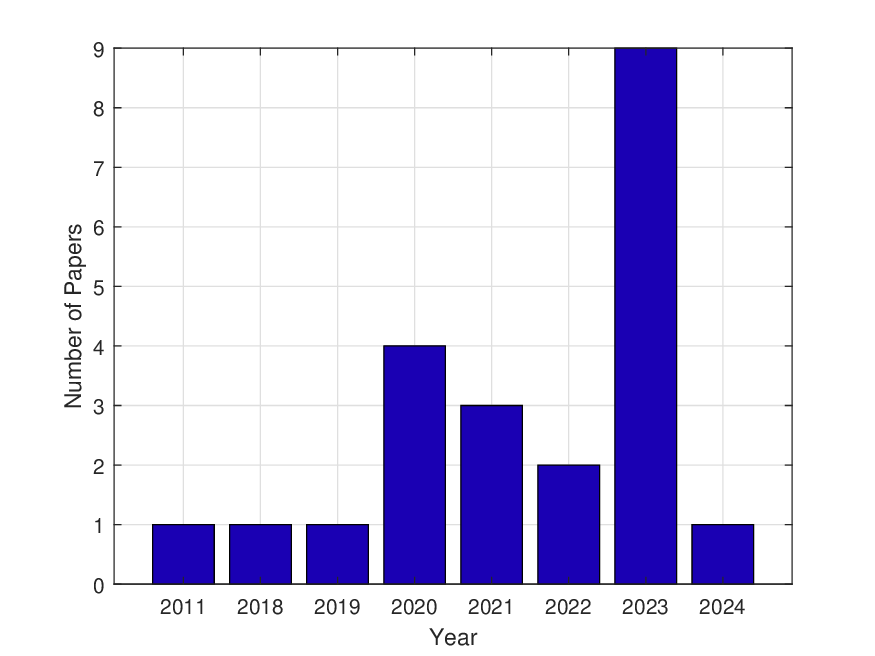}
    \caption{Distribution of studies by the year of publication}
    \label{fig:barchart}
\end{figure}
Figure \ref{fig:barchart} presents the distribution of studies by the year of publication.
While the earliest work is from 2011, there is a clear upward trend in the previous years, with a peak of 10 studies in 2023. 
Note that there was only one study in 2024 up until 5 March 2024, and more studies will likely appear throughout the remainder of the year. 
Overall, the increase from earlier years highlights the growing attention and research efforts in NMHE.


\begin{figure}[t]
    \centering
    \includegraphics[trim={13cm, 1.5cm, 0cm, 1cm}, clip,width=0.9\linewidth]{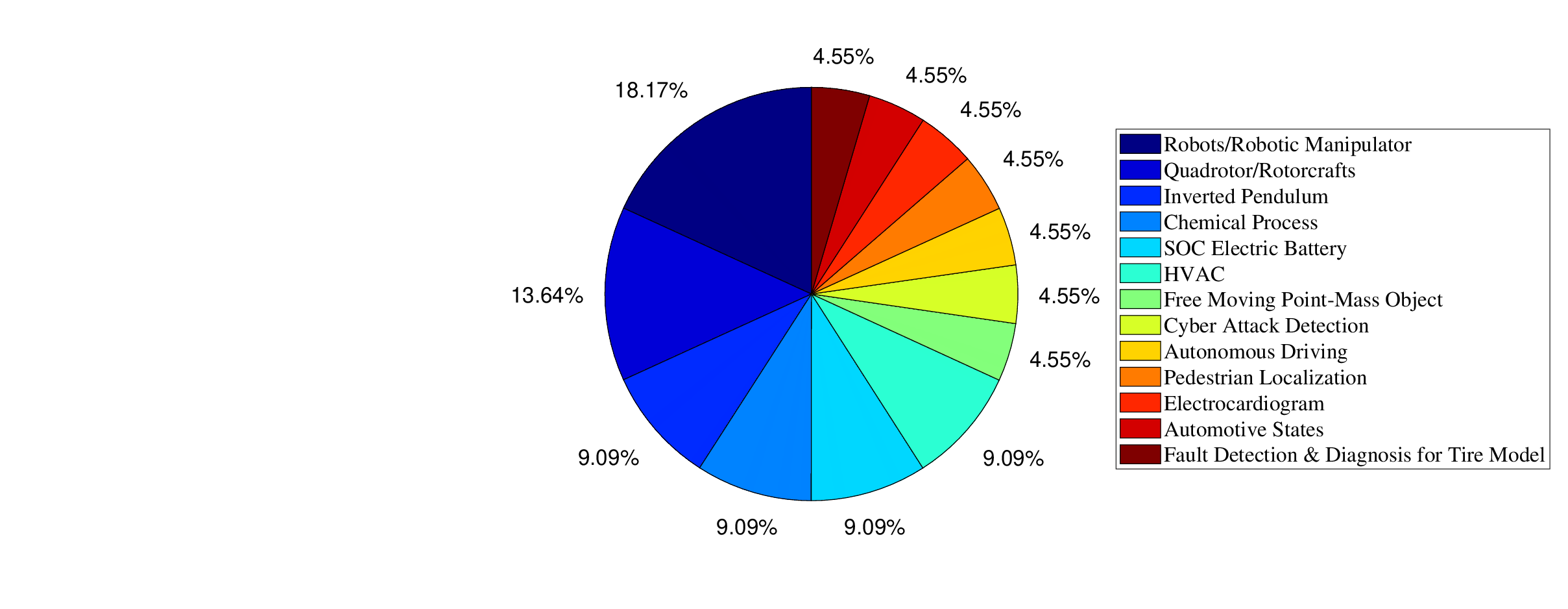}
    \caption{Distribution of studies by their application areas}
    \label{fig:piechart_application}
\end{figure}

Figure \ref{fig:piechart_application} illustrates a pie chart depicting the distribution of studies by their application areas. 
The chart reveals a broad range of applications, from systems with slower dynamics like heating, ventilation, and air conditioning (HVAC) to faster dynamics like rotorcraft.
Notably, about 40\% of the studies focus on rotorcraft, inverted pendulums, and robotics, all of which have highly nonlinear dynamics and pose significant state estimation and control challenges.
Additionally, many studies involve processes where an accurate system model is difficult to obtain, such as rotorcrafts due to the unmodeled aerodynamic effects of rotors, and pedestrian localization due to behavioral variability, environmental factors, and limited sensory information.
These observations underscore the particular value and effectiveness of NMHE in complex problems.

\begin{figure}
    \centering
    \includegraphics[trim={13cm, 2.5cm, 0cm, 1cm}, clip,width=0.8\linewidth]{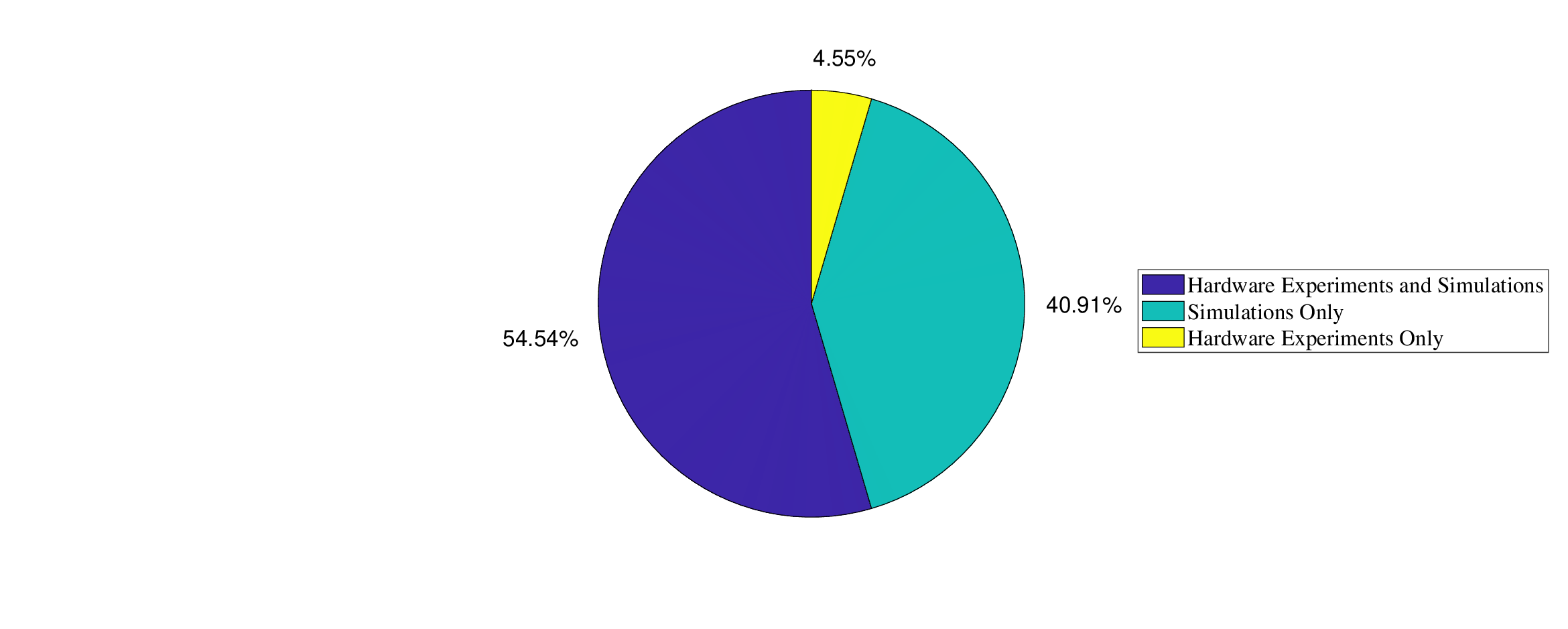}
    \caption{Distribution of studies based on the Method of NN-based MHE Implementation}
    \label{fig:piechart_sim_vs_exp}
\end{figure}

Hardware experimentation with NMHE can be challenging due to the complex structure of NNs and the need for nonlinear optimization solvers.
As such, we analyzed the distribution of studies that extended beyond simulation to include hardware experiments. 
Figure \ref{fig:piechart_sim_vs_exp} illustrates the results. 
Notably, 56.52\% of the studies implemented NMHE in real-time and conducted hardware experiments.
This observation demonstrates that with advancements in software tools and computing hardware, it has become feasible to implement NMHE in real-world scenarios.

\section{Different NMHE approaches}\label{se:narrative}
Table \ref{tab:all_studies} provides an overview of all studies included in this systematic review.
Based on the methodologies presented in these studies, we categorize existing NMHE work into three groups, as follows.
\begin{enumerate}
    \item The first group adopts the standard MHE formulation \eqref{eq:mhe_formulation}, but leverages NNs to create a more accurate model of the system, thereby enhancing state estimation accuracy.
    \item The second group employs NNs to modify the cost function \eqref{eq:cost}, providing auto-tuning capabilities that make the estimator adaptive to varying conditions.
    \item The third group uses NNs to approximate the standard MHE and implements the NN in place of the MHE for state estimation. This approach eliminates the need to solve the nonlinear constrained optimization problem inherent in MHE, leading to significant speedups, though it results in slightly lower state estimation accuracy since the NN is an approximation of the MHE.
\end{enumerate}
\begin{figure}[htbp]
    \centering
    \begin{subfigure}[b]{0.6\linewidth}
        \centering
        \includegraphics[width=\linewidth]{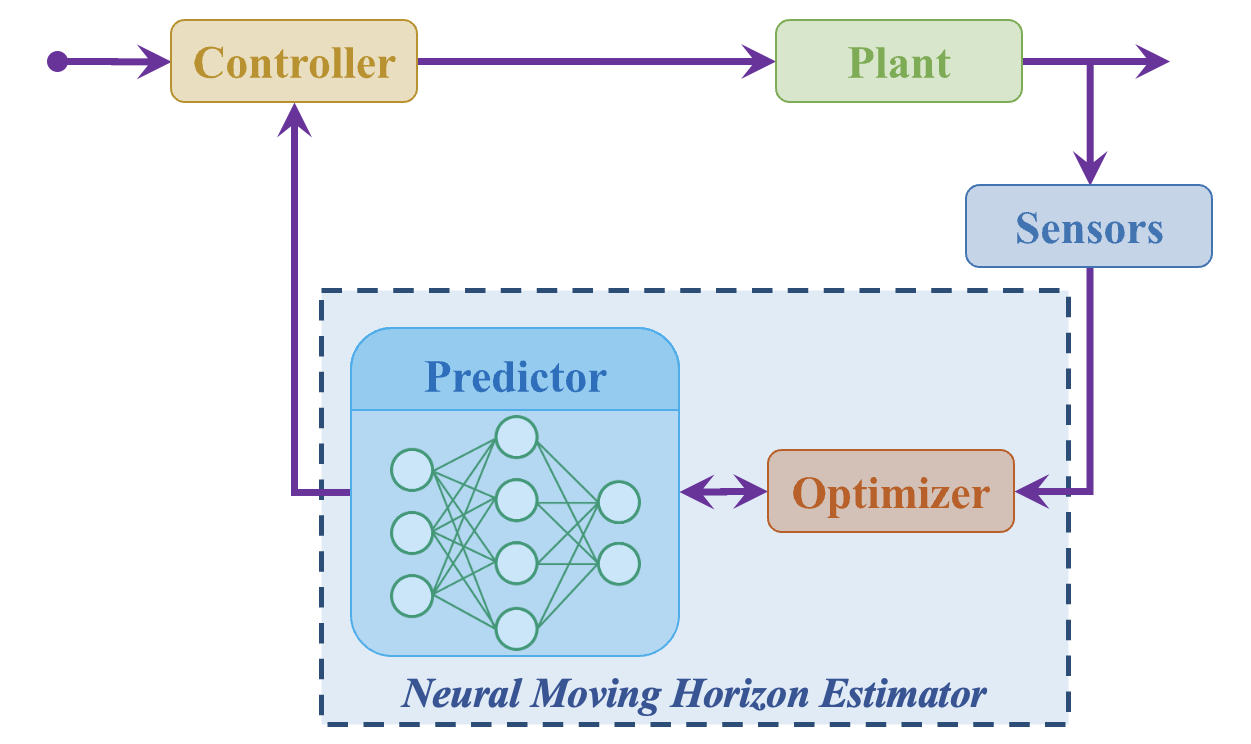} 
        \caption{Category I: using NNs for more accurate models}
        \label{fig:subfig1}
    \end{subfigure}
    \vfill
    \begin{subfigure}[b]{0.6\linewidth}
        \centering
        \includegraphics[width=\linewidth]{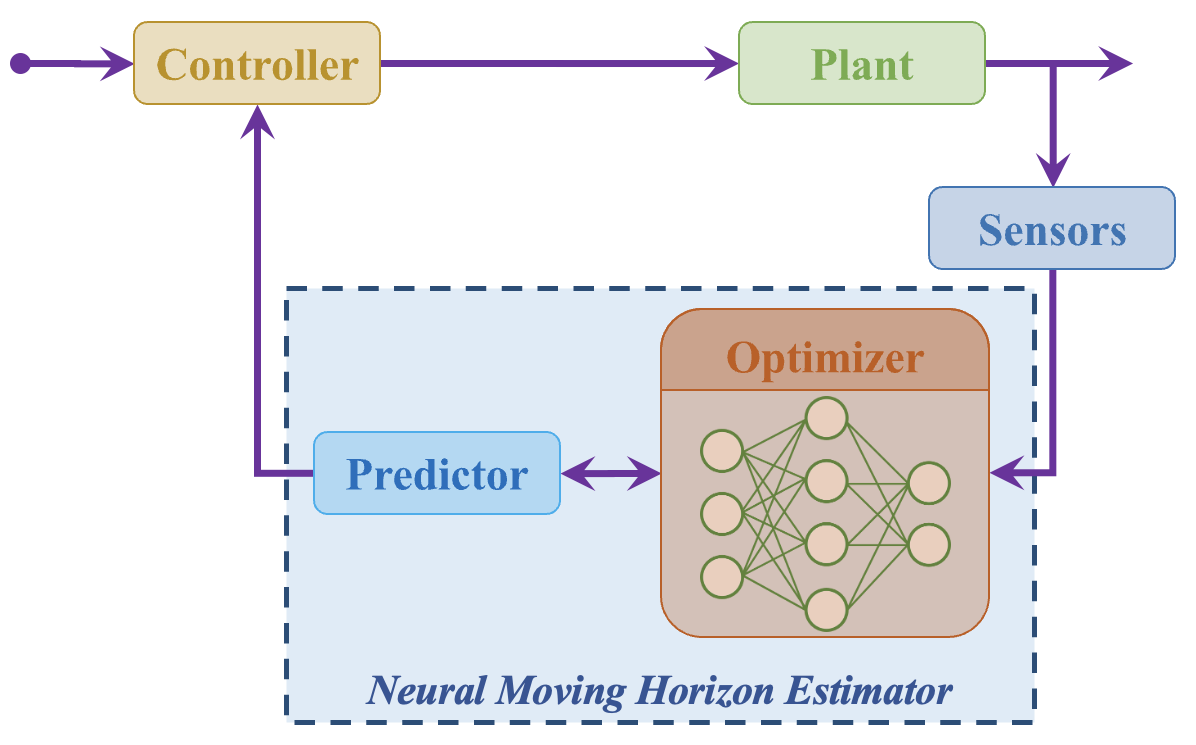} 
        \caption{Category II: using NNs to modify the cost function}
        \label{fig:subfig2}
    \end{subfigure}
    \vfill
    \begin{subfigure}[b]{0.6\linewidth}
        \centering
        \includegraphics[width=\linewidth]{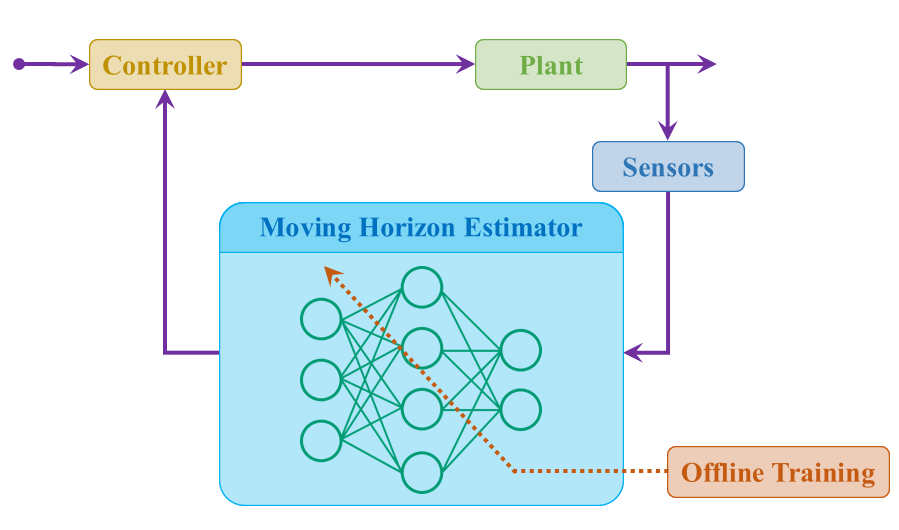} 
        \caption{Category III: using NNs for approximating regular MHE }
        \label{fig:subfig3}
    \end{subfigure}
    \caption{Different NMHE architectures in the existing studies}
    \label{fig:nmhe_configs}
\end{figure}

Figure \ref{fig:nmhe_configs} illustrates the architecture of these different approaches to NMHE. 
The rest of this section provides further details for each category.

\subsection{Category I: using NNs for more accurate models}
This category of studies builds upon the seminal work \cite{11}.
This study established theoretical results indicating that if the system model is represented by universal nonlinear approximators such as NNs, the MHE state estimations can converge to the actual state values, under certain conditions.
A large body of literature has embraced these theoretical findings, developing NMHE approaches for a broad spectrum of applications.
Comparative analyses across these studies consistently demonstrate the superior estimation accuracy of NMHE over traditional methods, irrespective of system complexity and computational demands.
Notable examples include \cite{7, 6, 3, 5, 23, 17, 15, 20}. 

The improved estimation accuracy provided by NMHE reduces the need for multiple and expensive sensors, enabling the development of more cost-effective and compact systems. 
This advantage is particularly evident in \cite{14}, which employs NMHE for estimating several parameters in brewery wastewater treatment, thereby eliminating the need for expensive sensors to measure each parameter.

Another key advantage of NMHE is the ability to explicitly account for system constraints while using an NN for accurate system model representation.
This capability is highlighted in \cite{4}, where NMHE is applied for parameter estimation in grey-box system identification.
Considering the known physical limitations of systems is proven highly effective in both offline and online parameter estimation.

NMHE excels in fault estimation and compensation. 
Concerning actuator faults, the study \cite{24} demonstrates the application of NMHE in estimating the current status of car tires to facilitate fault-tolerant chassis control. 
Regarding sensor faults, the study \cite{1} explores the use of NMHE in controlling complex process networks where cyberattacks can compromise the integrity of monitoring systems.
Due to the vast size and complexity of these networks, such attacks may go undetected, compromising system safety.
However, NMHE yields outstanding results in identifying these issues, recovering system states, and ensuring that control systems remain resilient to sensor faults and cyberattacks.


\subsection{Category II: using NNs to modify the cost function}
The MHE performance depends heavily on the choice of the weighting matrices $P$, $Q$, and $R$ in \eqref{eq:cost}.
Due to varying operating points and external disturbances, the optimal choice of these matrices is non-stationary, and finding appropriate values requires extensive data, making the MHE calibration process laborious.

Several papers have addressed this challenge. For example, the study \cite{8} replaces the cost function $J$ in \eqref{eq:mhe_formulation} with an NN.
This NN is trained online using a reinforcement learning algorithm, allowing $J$ to adapt over time to new operating points and external disturbances.

Another approach, named \textit{differential MHE} \cite{19}, updates the weighting matrices via gradient descent, with the gradient calculated recursively using a Kalman filter. In \cite{21}, differential MHE is extended to \textit{NeuroMHE}, where NN represents the weighting matrices.
The comparative analysis presented in \cite{21} demonstrates NeuroMHE's efficient training, fast adaptation, and improved estimation over the state-of-the-art estimators.

NeuroMHE is further refined in \cite{9} with a trust-region policy optimization method replacing gradient descent for tuning the weighting matrices. This method offers linear computational complexity with respect to the MHE horizon and requires minimal data and training time. In a case study on disturbance estimation for a quadrotor, the trust-region NeuroMHE achieved highly efficient training in under 5 minutes using only 100 data points, outperforming its gradient descent counterpart by up to 68.1\% in disturbance estimation accuracy while utilizing only 1.4\% of the network parameters used by the gradient descent method.

\subsection{Category III: using NNs for approximating regular MHE}
When the system model is nonlinear, MHE-based state estimation requires solving a constrained nonlinear optimization problem in real-time, which is challenging for resource-constrained systems. Several studies \cite{10, 12, 16, 13} have shown that an NN can closely resemble the behavior of MHE and replace it in the control system. Although the NN serves as an approximation of the MHE and may not achieve the same level of estimation accuracy, its parallel architecture aligns well with the multi-core architecture of modern processors. This compatibility makes NNs significantly faster than the constrained nonlinear optimization problems of MHE. Consequently, substantial speedups can be achieved at the expense of a slight reduction in state estimation accuracy.

Interestingly, in a case study on the state of charge (SOC) estimation of lithium-ion batteries \cite{13}, such an implementation of NMHE resulted in more accurate state estimation compared to the extended Kalman filter (EKF), while being over 98\% faster (0.1145 ms running time vs. 6.9789 ms). This demonstrates the significant potential of this class of NMHE approaches to deliver MHE-like state estimation accuracy with a fraction of the computational load of simple state estimators like EKF.

Such improvements in computational time open new doors to achieving MHE-like state estimation accuracy for complex systems that were previously challenging to address with the regular MHE. For example, the study \cite{12} develops an NMHE for an industrial batch polymerization reactor, a highly nonlinear system with 10 state variables, where solving a regular MHE can be difficult.
Furthermore, the study \cite{12} capitalizes on the computational efficiency of NNs to implement advanced control techniques like MPC alongside NMHE in a single control loop. 
Implementing two nonlinear constrained optimization problems in the same control loop in real-time may sound improbable or require substantial computational hardware, but with NN-based approximations, this becomes feasible on commodity hardware.

Studies \cite{10, 16} explore the implementation of NMHE on field-programmable gate arrays (FPGAs), achieving an impressive 500 kHz frame rate for state estimation of an inverted pendulum while maintaining MHE-like state estimation accuracy. This demonstrates the potential of this class of NMHE approaches to handle high-speed, real-time state estimation tasks effectively, faster and more accurately than common nonlinear state estimators like EKF, as mentioned above.

\section{NN architectures used in NMHE}\label{se:ml_details}
Unfortunately, not all studies provide the details of the NNs used in their NMHE developments. 
However, those that do reveal a variety of architectures, each customized to meet the unique needs of specific applications.
This section delves into the details of the structure of NNs used in these studies to gain deeper insights, beneficial for future NMHE designs.
We use the same categorization of studies presented in Section \ref{se:narrative} here.

\subsection{Category I: using NNs for more accurate models}
This category of studies commonly uses the multi-layer perceptron (MLP) networks to model system dynamics. Examples include \cite{1,4,6,17}.

The study \cite{17}, focusing on HVAC system state estimation, uses an MLP network with one hidden layer and three neurons.
The study \cite{4} employs an MLP network with one hidden layer and six neurons for grey-box system identification of a two-degrees-of-freedom (2-DOF) robotic manipulator.
The study \cite{6} investigates three case studies: (1) state estimation for a cartpole system using partial measurements, (2) localization for a ground robot, and (3) state estimation for a quadrotor. For the case study (1), an MLP network with 2 layers and 8 hidden neurons is trained over 250 epochs. For the case study (2), the same architecture is used but trained over 10,000 epochs. For the case study (3), an MLP network with 2 layers and 16 hidden neurons is trained over 150 epochs. 
The study \cite{1}, employs an MLP with two hidden layers for state estimation of complex network processes in the presence of cyberattacks.
The first hidden layer consists of 30 nodes, while the second has 24 nodes.
The input layer comprises tracking errors, while the output layer provides information on the presence or type of cyberattack.

In addition to MLP networks, long-term short memory (LSTM) and fuzzy networks have appeared in a few studies. 
The study \cite{3}, focusing on reliable state estimation of sideslip and attitude angles of cars, employs an LSTM network to model the errors of the inertial navigation system. This information is then used in an MHE to estimate the car side slip angle.
The network contains 20 hidden layers and 150 neurons.
The study \cite{23}, focusing on parameter identification and SOC estimation of lithium-ion batteries, utilizes an auto-regressive LSTM network with one hidden layer and 128 neurons. 
The study \cite{5}, focusing on car velocity and chassis control, implements an adaptive neuro-fuzzy inference system (ANFIS). 
The rule base of the network comprises 32 rules. 
Each input to the network is fuzzified using generalized Gaussian bell membership functions. 
The study \cite{7}, focusing on external force/torque estimation in bilateral telerobotics, uses type-2 fuzzy neural network (T2FNN). 
For a case study of 3-DOF master and 6-DOF slave manipulators, a T2FNN with 15 rules is proven effective.

Overall, we did not find clear justifications in the existing studies for why one type of network is favored over another. However, we have drawn the following observations:
\begin{itemize}
    \item MLP networks have simple yet effective structures. Among the studies that have employed MLP networks, one hidden layer seems to be sufficient for most cases. The number of neurons depends on the system's complexity, ranging from three neurons for a simple system like HVAC to 30 neurons for a complex network process. Additional estimation tasks, such as identifying the type of cyberattacks, appear to require more hidden layers.
    \item LSTM networks, capable of capturing temporal dependencies, excel in dynamical systems modeling. However, there was only two studies that employed LSTM, featuring a much more complex structure compared to the aforementioned MLP networks. Unfortunately, there is no direct comparison of LSTM and MLP networks in the context of NMHE, making it difficult to draw definitive conclusions. However, given the known advantages of LSTM networks and the outstanding state estimation results obtained with them in \cite{3}, LSTMs are worth considering, especially for high-performance applications.
    \item The results obtained with fuzzy networks are noteworthy. Compared to MLP networks, the design of fuzzy networks is a more involved process as they require a degree of domain knowledge to set up the membership functions and the rule base. However, this initial setup has the potential to reduce the amount of time required for training. Considering that MLP training reached up to 10,000 epochs in the aforementioned studies, the potential of fuzzy networks to reduce training times makes them an attractive option.
\end{itemize}



\subsection{Category II: using NNs to modify the cost function}
In \cite{8}, an NN is used to approximate the cost function. Convexity of the cost function is often desired because it guarantees that any local minimum is also a global minimum, simplifying the optimization process. It also facilitates the use of robust and efficient convex optimization solvers. Consequently, the study \cite{8} employs an input convex neural network (ICNN) to approximate the cost function. This ICNN consists of two hidden layers, each containing 26 neurons. This configuration is applied for state estimation in a building heat control system.

In contrast to \cite{8}, the rest of the studies in this category \cite{19, 21, 9} do not use NNs to approximate the entire cost function, but rather just the weighting matrices. Consequently, the convexity of NNs is no longer required, allowing for the use of conventional architectures such as MLP networks. For instance, in NeuroMHE \cite{21}, the MLP network consists of two hidden layers, each with 50 neurons. In \cite{9}, the MLP network is composed of two hidden layers, each with eight neurons. Although this is a simpler NN compared to the one in \cite{19}, it performs better due to superior training algorithms.

The key observations in this category include:
\begin{itemize}
    \item If an NN is used to approximate the entire cost function, the use of networks that guarantee convexity, e.g., ICNN, is desired.
    \item In the NeuroMHE setup, where an NN is used to approximate the weighing matrices, MLP networks are effective; however, the use of a trust-region optimization algorithm helps simplify the network structure while offering higher state estimation accuracy. 
\end{itemize}

\subsection{Category III: using NNs for approximating regular MHE}
As explained in Section \ref{se:narrative}, this category of studies is distinct in the sense that it focuses on approximating the regular MHE to achieve faster running times, rather than improving the estimation accuracy.

The study \cite{13} develops a regular MHE for the state-of-charge estimation of lithium-ion batteries, subsequently approximated using an MLP network with 20 neurons and 5 hidden layers. The NN closely mimics the MHE, achieving a coefficient of determination near 99\%. Remarkably, the network operates 20 times faster than the regular MHE.
The reported execution time for the network is 0.1145 [ms], which is even faster than the reported EKF execution time of 6.9789 [ms].  This substantial speedup, coupled with only a 1\% reduction in accuracy, underscores the potential of NN approximations for MHE, particularly in resource-constrained environments.

The approximating network enlarges as the complexity of the system increases.
The study \cite{12} explores a regular MHE for an industrial batch polymerization reactor, a significantly more complex system compared to the aforementioned lithium-ion battery case study. 
The study uses an MLP network with 109 inputs, 80 neurons per hidden layer, 6 hidden layers, and 9 outputs to approximate the MHE. 
The 109 inputs correspond to all past measurements required for MHE calculations, resulting in a highly complex network.
As mentioned in Section \ref{se:narrative}, this study also investigates the integration of NMHE with MPC, approximating MPC with another NN.
The NN mimicking the MPC has 9 inputs, 25 neurons per hidden layer, 6 hidden layers, and 93 outputs, revealing that approximating MHE necessitates much larger networks than MPC, likely due to the greater number of inputs needed for meaningful state estimation.
Further, this study trains an NN to approximate the control system that includes both MHE and MPC, employing an MLP with 109 inputs, 120 neurons per hidden layer, 8 hidden layers, and 12 outputs.

The studies \cite{10, 16} present a fast NMHE specifically designed for an inverted pendulum system.
Initially, an MHE is crafted for the system, followed by the application of radial basis function (RBF) networks to approximate the developed MHE.
The study experiments with networks featuring a single hidden layer and a varying number of neurons (ranging from 10 to 16).
It concludes that an RBF network with 14 neurons is sufficient to appropriately approximate the regular MHE.
Implementing this network on FPGAs achieves an impressive 500 kHz framerate.
Comparative analysis with another embedded computing platform i.e. the Raspberry Pi 3, reveals that the FPGA implementation is 31 times faster.

The main takeaways from this category are:
\begin{itemize}
    \item The MLP networks used in this category are considerably larger than the ones used in the previous categories, as the NN represents the entire state estimator, including the system model, and the optimizer. In the previous categories, the NN represented only the system model or a portion of the optimizer.
    \item The RBF network developed in \cite{10, 16} appears to be much smaller than the MLP networks used in \cite{13, 12}. While these two types of networks have not been compared directly in one case study, the simplicity of the RBF network makes it worth considering for future applications.
\end{itemize}

\section{NMHE running time}\label{se:running_time}
Apart from studies that utilize NNs to approximate regular MHEs for faster implementations, most NMHE approaches remain computationally intensive. This is due to the simultaneous need to run NN models and nonlinear constrained optimization algorithms. Unfortunately, not all studies provide details of their NMHE running times and computing hardware. For those that do, Table \ref{tab:comptime} compiles the running times of NMHE implementations, along with the computing hardware and the length of the horizon, both of which significantly impact the computational load of the estimator.

It is evident that, for the majority of studies, the NMHE running time is only a few milliseconds. However, this performance is typically achieved on general-purpose processors. While this may not pose an issue for applications like process control, where powerful computers may be available, it is impractical for mobile applications, especially in quadrotors. In such cases, embedded processors are necessary, and it remains unclear how fast NMHE will perform on these platforms.

With recent advances in parallel computing and the use of fast optimization solvers like ACADOS \cite{verschueren2022acados}, the real-time application of NMHE might be feasible. This is a topic that warrants further exploration in the near future.

One definitive conclusion regarding hardware implementations is that NN-based approximation techniques significantly speed up state estimations. They offer much more efficient computations compared to simple algorithms like extended Kalman filter, while also bringing the benefits of MHE to state estimation.

\renewcommand{\arraystretch}{1.5} 
\begin{table}[ht]
\begin{scriptsize}
\caption{NMHE running in hardware experimentation and relevant factors}
\label{tab:comptime}
\centering
\begin{tabular}{@{}p{1cm}p{5.5cm}p{2cm}p{2cm}p{3.5cm}@{}}
\toprule  
Reference & Application area & Running time & Horizon length & Computing hardware\\
\midrule
\cite{11} & Quadrotors & 1.66 [ms] & 4 & Intel Xeon\\
\cite{5} & Self-driving car & 4.8 [ms] & 6 & Intel Core i7-8550U\\
\cite{15} & Pedestrian localization & 55 [ms] & - & Intel Core i7-9750\\
\cite{20} & Electrocardiogram & 801 [$\mu$s] & 5 & Raspberry Pi\\
\cite{14} & SOC estimation of lithium-ion batteries & 0.08 [s] & 20 & Intel Core i7\\
\cite{24} & Fault detection \& diagnosis for car tire & 11.28 [s] & 10 & Intel Core i5-8400\\
\cite{19} & Quadrotors & 5.6 [ms] & 20 & AMD Ryzen 9 5950X\\
\cite{21} & Quadrotors & 1.83 [ms] & 10 & Intel Core i7-11700K\\
\cite{9} & Quadrotors & 1.83 [ms] & 10 & Intel Core i7-11700K\\
\cite{10} & Inverted pendulum & 1.60 [$\rm{\mu}$s] & 20 & FPGA\\
\cite{12} & Industrial batch polymerization reactor & 191 [ms] & 20 & ARM Cortex-M0+\\
\cite{16} & Inverted pendulum & 1.60 [$\rm{\mu}$s] & 9 & FPGA\\
\cite{13} & SOC estimation of lithium-ion batteries & 0.1146 [ms] & 20 & Nvidia Jetson Nano \\
\bottomrule
\end{tabular}
\end{scriptsize}
\end{table}

\section{Future Directions}
NMHE is still in its infancy and requires extensive research. This section presents potential future directions to advance NMHE.

One important direction involves training NN models. A critical challenge in leveraging NNs for NMHE is generating high-quality datasets. An emerging field in system modeling is physics-informed NNs (PINNs), which can incorporate physical laws and constraints into the training process. This approach leads to more accurate and reliable models while generally requiring less data. Integrating PINNs with MHE holds the potential to reduce the need for large training datasets while simultaneously improving NMHE state estimation accuracy.

As discussed in Section \ref{se:ml_details}, most NNs used in existing studies are MLP networks. However, recurrent NNs (RNNs), including LSTM networks, are known for their ability to detect time dependencies and model dynamical systems effectively. Surprisingly, we found only two studies that utilized RNNs \cite{3, 23}. Therefore, future research should further explore the applications of RNNs.

In addition to RNNs, convolutional NNs (CNNs) and transformer networks have also demonstrated exceptional performance in time series prediction and can be adapted for NMHE. Moreover, fuzzy networks and RBF networks have also shown impressive results in the existing studies and should not be excluded from future work.

There has been little work on stochastic formulations for NMHE. 
Given the presence of process and measurement noise, advanced stochastic optimization techniques, such as Monte Carlo methods and Bayesian approaches, can significantly improve the performance of NMHE in the presence of stochastic uncertainties.

To translate NMHE into practice, more work should focus on hardware implementations. As mentioned in Section \ref{se:running_time}, it is unclear how NMHE implementations perform on embedded processors.
Focusing on efficient implementations of NNs in parallel processors, and using numerically efficient solvers, will optimize the computational efficiency of NMHE and enable real-world hardware experiments.
\section{Conclusion}
In this systematic review, we have explored the current state and future directions of NMHE in state estimation. There are three distinct categories of approaches, each offering significant advantages in state estimation accuracy or computational efficiency.
We identified several key areas for future research to advance NMHE. Existing studies on NMHE have not fully leveraged the latest advances in NNs. The use of PINNs, RNNs, CNNs, and transformer networks in NMHE is underexplored. The use of such advanced NNs can potentially improve the estimation accuracy and computational efficiency of NMHE. In addition, exploring stochastic formulations can better account for process and measurement noise, enhancing the robustness of state estimation.
Moreover, leveraging parallel computing and fast optimization solvers is critical to reducing the computing time of NMHE and implementing it in embedded hardware platforms for real-world applications.
By addressing these research directions, NMHE can become a more robust, efficient, and versatile tool for state estimation, expanding its applicability across various industries.

\bibliographystyle{elsarticle-num} 
\bibliography{References}

\begin{thebibliography}{10}
\expandafter\ifx\csname url\endcsname\relax
  \def\url#1{\texttt{#1}}\fi
\expandafter\ifx\csname urlprefix\endcsname\relax\def\urlprefix{URL }\fi
\expandafter\ifx\csname href\endcsname\relax
  \def\href#1#2{#2} \def\path#1{#1}\fi

\bibitem{simon2006optimal}
D.~Simon, Optimal state estimation: Kalman, H infinity, and nonlinear approaches, John Wiley \& Sons, 2006.

\bibitem{khodarahmi2023review}
M.~Khodarahmi, V.~Maihami, A review on kalman filter models, Archives of Computational Methods in Engineering 30~(1) (2023) 727--747.

\bibitem{doucet2009tutorial}
A.~Doucet, A.~M. Johansen, et~al., A tutorial on particle filtering and smoothing: Fifteen years later, Handbook of nonlinear filtering 12~(656-704) (2009) 3.

\bibitem{rawlings2021moving}
J.~B. Rawlings, D.~A. Allan, Moving horizon estimation, in: Encyclopedia of Systems and Control, Springer, 2021, pp. 1352--1358.

\bibitem{s21062085}
X.-B. Jin, R.~J. Robert~Jeremiah, T.-L. Su, Y.-T. Bai, J.-L. Kong, The new trend of state estimation: From model-driven to hybrid-driven methods, Sensors 21~(6) (2021).

\bibitem{feng2023review}
S.~Feng, X.~Li, S.~Zhang, Z.~Jian, H.~Duan, Z.~Wang, A review: state estimation based on hybrid models of kalman filter and neural network, Systems Science \& Control Engineering 11~(1) (2023) 2173682.

\bibitem{bai2023state}
Y.~Bai, B.~Yan, C.~Zhou, T.~Su, X.~Jin, State of art on state estimation: Kalman filter driven by machine learning, Annual Reviews in Control 56 (2023) 100909.

\bibitem{inbook}
J.~Rawlings, Moving Horizon Estimation, 2013, pp. 1--7.
\newblock \href {https://doi.org/10.1007/978-1-4471-5102-9_4-1} {\path{doi:10.1007/978-1-4471-5102-9_4-1}}.

\bibitem{schwenzer2021review}
M.~Schwenzer, M.~Ay, T.~Bergs, D.~Abel, Review on model predictive control: An engineering perspective, The International Journal of Advanced Manufacturing Technology 117~(5) (2021) 1327--1349.

\bibitem{FAGIANO2013193}
L.~Fagiano, C.~Novara, \href{https://www.sciencedirect.com/science/article/pii/S0005109812004700}{A combined moving horizon and direct virtual sensor approach for constrained nonlinear estimation}, Automatica 49~(1) (2013) 193--199.
\newblock \href {https://doi.org/https://doi.org/10.1016/j.automatica.2012.09.009} {\path{doi:https://doi.org/10.1016/j.automatica.2012.09.009}}.
\newline\urlprefix\url{https://www.sciencedirect.com/science/article/pii/S0005109812004700}

\bibitem{wan2018real}
Y.~Wan, T.~Keviczky, Real-time fault-tolerant moving horizon air data estimation for the reconfigure benchmark, IEEE Transactions on Control Systems Technology 27~(3) (2018) 997--1011.

\bibitem{tenny2002efficient}
M.~J. Tenny, J.~B. Rawlings, Efficient moving horizon estimation and nonlinear model predictive control, in: Proceedings of the 2002 American Control Conference (IEEE Cat. No. CH37301), Vol.~6, IEEE, 2002, pp. 4475--4480.

\bibitem{bae2017humanoid}
H.~Bae, J.-H. Oh, Humanoid state estimation using a moving horizon estimator, Advanced Robotics 31~(13) (2017) 695--705.

\bibitem{zanon2013nonlinear}
M.~Zanon, J.~V. Frasch, M.~Diehl, Nonlinear moving horizon estimation for combined state and friction coefficient estimation in autonomous driving, in: 2013 European control conference (ECC), IEEE, 2013, pp. 4130--4135.

\bibitem{vandersteen2013spacecraft}
J.~Vandersteen, M.~Diehl, C.~Aerts, J.~Swevers, Spacecraft attitude estimation and sensor calibration using moving horizon estimation, Journal of Guidance, Control, and Dynamics 36~(3) (2013) 734--742.

\bibitem{kraus2013moving}
T.~Kraus, H.~J. Ferreau, E.~Kayacan, H.~Ramon, J.~De~Baerdemaeker, M.~Diehl, W.~Saeys, Moving horizon estimation and nonlinear model predictive control for autonomous agricultural vehicles, Computers and electronics in agriculture 98 (2013) 25--33.

\bibitem{11}
A.~Alessandri, M.~Baglietto, G.~Battistelli, M.~Gaggero, Moving-horizon state estimation for nonlinear systems using neural networks, IEEE Transactions on Neural Networks 22~(5) (2011) 768--780.

\bibitem{7159360}
K.~J. Nidhil~Wilfred, S.~Sreeraj, B.~Vijay, V.~Bagyaveereswaran, System identification using artificial neural network, in: 2015 International Conference on Circuits, Power and Computing Technologies [ICCPCT-2015], 2015, pp. 1--4.
\newblock \href {https://doi.org/10.1109/ICCPCT.2015.7159360} {\path{doi:10.1109/ICCPCT.2015.7159360}}.

\bibitem{2}
L.~Ecker, M.~Sch{\"o}berl, Data-driven observer design for an inertia wheel pendulum with static friction, IFAC-PapersOnLine 55~(40) (2022) 193--198.

\bibitem{22}
W.~Choo, E.~Kayacan, Data-based mhe for agile quadrotor flight, in: 2023 IEEE/RSJ International Conference on Intelligent Robots and Systems (IROS), IEEE, 2023, pp. 4307--4314.

\bibitem{page2021prisma}
M.~J. Page, J.~E. McKenzie, P.~M. Bossuyt, et~al., \href{https://doi.org/10.1186/s13643-021-01626-4}{The prisma 2020 statement: an updated guideline for reporting systematic reviews}, Systematic Reviews 10~(1) (2021) 89.
\newblock \href {https://doi.org/10.1186/s13643-021-01626-4} {\path{doi:10.1186/s13643-021-01626-4}}.
\newline\urlprefix\url{https://doi.org/10.1186/s13643-021-01626-4}

\bibitem{7}
D.~Sun, Q.~Liao, T.~Stoyanov, A.~Kiselev, A.~Loutfi, Bilateral telerobotic system using type-2 fuzzy neural network based moving horizon estimation force observer for enhancement of environmental force compliance and human perception, Automatica 106 (2019) 358--373.

\bibitem{6}
K.~Y. Chee, M.~A. Hsieh, Learnest: Learning enhanced model-based state estimation for robots using knowledge-based neural ordinary differential equations, in: 2023 IEEE International Conference on Robotics and Automation (ICRA), IEEE, 2023, pp. 11590--11596.

\bibitem{3}
R.~Song, Y.~Fang, H.~Huang, Reliable estimation of automotive states based on optimized neural networks and moving horizon estimator, IEEE/ASME Transactions on Mechatronics (2023).

\bibitem{5}
E.~Alcala, O.~Sename, V.~Puig, J.~Quevedo, Ts-mpc for autonomous vehicle using a learning approach, IFAC-PapersOnLine 53~(2) (2020) 15110--15115.

\bibitem{23}
Y.~Chen, C.~Li, S.~Chen, H.~Ren, Z.~Gao, A combined robust approach based on auto-regressive long short-term memory network and moving horizon estimation for state-of-charge estimation of lithium-ion batteries, International Journal of Energy Research 45~(9) (2021) 12838--12853.

\bibitem{17}
S.~Mostafavi, H.~Doddi, K.~Kalyanam, D.~Schwartz, Nonlinear moving horizon estimation and model predictive control for buildings with unknown hvac dynamics, IFAC-PapersOnLine 55~(41) (2022) 71--76.

\bibitem{15}
E.~Mohammadbagher, N.~P. Bhatt, E.~Hashemi, B.~Fidan, A.~Khajepour, Real-time pedestrian localization and state estimation using moving horizon estimation, in: 2020 IEEE 23rd International Conference on Intelligent Transportation Systems (ITSC), IEEE, 2020, pp. 1--7.

\bibitem{20}
S.~Banerjee, G.~K. Singh, A new moving horizon estimation based real-time motion artifact removal from wavelet subbands of ecg signal using particle filter, Journal of Signal Processing Systems 95~(8) (2023) 1021--1035.

\bibitem{14}
L.~Dewasme, Neural network-based software sensors for the estimation of key components in brewery wastewater anaerobic digester: an experimental validation, Water Science and Technology 80~(10) (2019) 1975--1985.

\bibitem{4}
K.~F. L{\o}wenstein, D.~Bernardini, L.~Fagiano, A.~Bemporad, Physics-informed online learning of gray-box models by moving horizon estimation, European Journal of Control 74 (2023) 100861.

\bibitem{24}
B.~Zhang, S.~Lu, W.~Xie, F.~Xie, Fault detection and diagnosis based on interactive multi-model moving horizon estimation and neuro-tire model, IEEE/ASME Transactions on Mechatronics (2024).

\bibitem{1}
B.~Sundberg, D.~B. Pourkargar, Cyberattack awareness and resiliency of integrated moving horizon estimation and model predictive control of complex process networks, in: 2023 American Control Conference (ACC), IEEE, 2023, pp. 3815--3820.

\bibitem{8}
H.~N. Esfahani, A.~B. Kordabad, W.~Cai, S.~Gros, Learning-based state estimation and control using mhe and mpc schemes with imperfect models, European Journal of Control 73 (2023) 100880.

\bibitem{19}
B.~Wang, Z.~Ma, S.~Lai, L.~Zhao, T.~H. Lee, Differentiable moving horizon estimation for robust flight control, in: 2021 60th IEEE Conference on Decision and Control (CDC), IEEE, 2021, pp. 3563--3568.

\bibitem{21}
B.~Wang, Z.~Ma, S.~Lai, L.~Zhao, Neural moving horizon estimation for robust flight control, IEEE Transactions on Robotics (2023).

\bibitem{9}
B.~Wang, X.~Chen, L.~Zhao, Trust-region neural moving horizon estimation for robots, arXiv preprint arXiv:2309.05955 (2023).

\bibitem{10}
R.~K.~V. Brunello, Nonlinear moving-horizon state estimation for hardware implementation and a model predictive control application (2021).

\bibitem{12}
B.~Karg, S.~Lucia, Approximate moving horizon estimation and robust nonlinear model predictive control via deep learning, Computers \& Chemical Engineering 148 (2021) 107266.

\bibitem{16}
R.~K.~V. Brunello, R.~C. Sampaio, C.~H. Llanos, L.~dos Santos~Coelho, H.~V.~H. Ayala, Efficient hardware implementation of nonlinear moving-horizon state estimation with artificial neural networks, IFAC-PapersOnLine 53~(2) (2020) 7813--7818.

\bibitem{13}
E.~D.~R. Lopes, M.~M. Soudre, C.~H. Llanos, H.~V.~H. Ayala, Nonlinear receding-horizon filter approximation with neural networks for fast state of charge estimation of lithium-ion batteries, Journal of Energy Storage 68 (2023) 107677.

\bibitem{verschueren2022acados}
R.~Verschueren, G.~Frison, D.~Kouzoupis, J.~Frey, N.~v. Duijkeren, A.~Zanelli, B.~Novoselnik, T.~Albin, R.~Quirynen, M.~Diehl, acados—a modular open-source framework for fast embedded optimal control, Mathematical Programming Computation 14~(1) (2022) 147--183.

\bibitem{18}
E.~Kayacan, Z.-Z. Zhang, G.~Chowdhary, Embedded high precision control and corn stand counting algorithms for an ultra-compact 3d printed field robot, in: Robotics: science and systems, Vol.~14, 2018, p.~9.

\end{thebibliography}
\newpage
\section*{Appendix A}
\renewcommand{\arraystretch}{1.5} 
\begin{scriptsize}
\begin{longtable}{@{}p{1cm} >{\raggedright\arraybackslash}p{6cm} >{\raggedright\arraybackslash}p{6cm}@{}}
\caption{List of NMHE studies included in this systematic review}
\label{tab:all_studies} \\
\toprule  
Reference & Objective & Case study \\
\midrule
\endfirsthead
\caption[]{Details of Studies Included (continued)} \\
\toprule
Reference & Objective & Application area \\
\midrule
\endhead
\bottomrule
\endfoot
\cite{11} & To use NNs as the predictor in MHE formulation, leveraging NNs for better state estimation accuracy & Free moving point-mass object \\ 
\cite{7} & To estimate external force/torque information and disturbance rejection using NMHE & Bilateral teleoperation of robotic manipulators \\ 
\cite{6} & To leverage NMHE for enhanced state estimation in different applications & State estimation for an inverted pendulum using partial measurements, localization for a ground robot, and state estimation for a quadrotor \\
\cite{3} & To enhance the accuracy of car sideslip angle estimation using NMHE & Car chassis control \\
\cite{5} & To leverage NMHE for high-performance state estimation & Autonomous racing of cars \\ 
\cite{23} & Identify battery parameters and estimate SOC with high precision & Lithium-ion batteries \\
\cite{17} & Estimate HVAC systems state variables using only building management system data & Building HVAC control for occupant comfort satisfaction and energy-savings\\ 
\cite{15} & For online estimation of pedestrian position, velocity, and acceleration & Pedestrian localization for car autonomous navigation\\ 
\cite{20} & To remove motion artifact from electrocardiogram signals & Electrocardiogram signal processing\\ 
\cite{14} & Estimate state variables using basic measurements for process monitoring, eliminating the need for expensive sensors & Online monitoring of brewery wastewater anaerobic digester \\ 
\cite{4} & Reduce modeling errors in gray-box system identification & 2-DOF robotic manipulator \\
\cite{24} & Overcome the low accuracy of traditional tire models, enabling high-fidelity fault identification & Fault detection and diagnosis of car tires\\ 
\cite{1} & Detect cyberattacks in complex process networks & The integrated process of benzene alkylation with ethylene to produce ethylbenzene \\ 
\cite{18} & Identify terrain parameters using low-cost sensors & Mobile robot \\ 
\cite{8} & Enable accurate state estimation in the absence of reliable system models & Climate control of smart building \\
\cite{19} & Develop an auto-tuning state estimator for varying conditions & Quadrotor disturbance estimation \\ 
\cite{21} & Develop an auto-tuning state estimator for varying conditions & Quadrotor disturbance estimation \\ 
\cite{9} & Use trust-region policy optimization for an adaptable state estimator & Quadrotor disturbance estimation \\
\cite{10} & Approximate MHE with NNs for fast state estimation & Inverted Pendulum \\ 
\cite{12} & Approximate MHE with NNs for fast state estimation, integrating it with MPC & Industrial batch polymerization reactor \\ 
\cite{16} & Approximate MHE with NNs for fast state estimation & Inverted Pendulum \\ 
\cite{13} & Approximate MHE with NNs for fast state estimation & SOC estimation of lithium-ion batteries \\ 
\end{longtable}
\end{scriptsize}

\end{document}